**[Article Full Title]:** A multimodal ensemble approach for clear cell renal cell carcinoma treatment outcome prediction

**[Short Running Title]:** Multimodal ensemble ccRCC outcome prediction

**[Author Names]:** Meixu Chen, PhD[1], Kai Wang, PhD[1,2], Payal Kapur, MD[3], James Brugarolas, MD, PhD[4], Raquibul Hannan, MD, PhD[1] and Jing Wang, PhD[1]

**[Author Institutions]:**
[1] *Medical Artificial Intelligence and Automation (MAIA) Lab, Department of Radiation Oncology, University of Texas Southwestern Medical Center, Dallas, TX, 75235, USA.*
[2] *Department of Radiation Oncology, University of Maryland Medical Center, Baltimore, MD 21201, USA.*
[3] *Department of Pathology and Urology, UT Southwestern Medical Center, Dallas, TX, 75235, USA.*
[4] *Department of Internal Medicine, UT Southwestern Medical Center, Dallas, TX, 75235, USA.*

**[Corresponding Author Name & Email Address]:**
Corresponding Author: Jing Wang, PhD
Email: Jing.Wang@UTSouthwestern.edu
*Medical Artificial Intelligence and Automation (MAIA) Lab, Department of Radiation Oncology, UT Southwestern Medical Center, Dallas, TX, 75235, USA.*

**[Author Responsible for Statistical Analysis Name & Email Address]**
Meixu Chen, PhD
Email: Meixu.Chen@UTSouthwestern.edu

**[Conflict of Interest Statement for All Authors]**
*Conflict of Interest: Nothing to disclose.*

**[Funding Statement]**
*This work is partially supported by NIH P50CA196516.*

**[Data Availability Statement for this Work]**
*The Kidney Renal Clear Cell Carcinoma (KIRC) dataset used in this study is publicly available from The Cancer Genome Atlas (TCGA) repository. It can be accessed via the Genomic Data Commons (GDC) Data Portal at https://portal.gdc.cancer.gov/. For additional information on how to access and download the TCGA-KIRC dataset, please refer to the following link: https://portal.gdc.cancer.gov/projects/TCGA-KIRC. Pre-processed clinical and multi-omics data are publicly available on many websites, including cBioPortal, Firebrowse, LinkedOmics, or TCGA Preprocessed Multi-Omics Cancer Benchmark Dataset.*

**Abstract**

**Purpose:** A reliable and comprehensive cancer prognosis model for clear cell renal cell carcinoma (ccRCC) could better assist in personalizing treatment. In this work, we developed a multi-modal ensemble model (MMEM) which integrates pretreatment clinical information, multi-omics data, and histopathology whole slide image (WSI) data to learn complementary information to predict overall survival (OS) and disease-free survival (DFS) for patients with ccRCC.

**Methods and Materials:** We collected 226 patients from The Cancer Genome Atlas Kidney Renal Clear Cell Carcinoma dataset (TCGA-KIRC). These patients have OS and DFS follow up data available and five different data modalities provided, including clinical information, pathology data in the form of WSI, and three multi-omics data, which comprise mRNA expression, miRNA expression (miRSeq), and DNA methylation data. Five sets of separate survival prediction models were constructed separately for OS and DFS. We used a traditional Cox-proportional hazards (CPH) model with iterative forward feature selection for clinical and multi-omics data. Four different types of pre-trained encoder models, comprising ResNet and three recently developed general purpose foundation models for computational pathology, were utilized to extract features from processed WSI patches. A deep learning-based CPH model was constructed to predict survival outcomes using these encoded WSI features. For each of the survival outcomes of interest, we weigh and combine the predicted risk scores from all the five models to generate the final prediction. Model weighting was based on the training performance. Five-fold cross validation was performed to train and test the proposed workflow.

**Results:** We employed the concordance index (C-index) and area under the receiver operating characteristic curve (AUROC) metrics to assess the performance of our models for time-to-event prediction and time-specific binary prediction, respectively. Among the sub-models, the clinical feature based CPH model has the highest weight for both prediction tasks. For WSI-based prediction, the encoded feature using an image-based general purpose foundation model (UNI) showed the best prediction performance over other pretrained feature encoders. Our final model outperformed corresponding single-modality models on all prediction labels, achieving C-indices of 0.820 and 0.833 for OS and DFS, respectively. The AUROC values for binary prediction at follow-up of 3 year were 0.831 and 0.862 for patient death and cancer recurrence, respectively. Using the medians of predicted risks as thresholds to identify high-risk and low-risk patient groups, we performed log-rank tests, which revealed improved performance in both OS and DFS compared to single-modality models.

**Conclusion:** We developed the first multi-modal prediction model MMEM for ccRCC patients that integrates features across five different data modalities. Our model demonstrated better prognostic ability compared with corresponding single-modality models for both prediction targets. If findings are independently reproduced, it has the potential to assist in management of ccRCC patients.

## Introduction

Kidney cancer, mostly renal cell carcinoma (RCC), is one of the most common cancers in the United States, and it is estimated to have about 81,610 new cases (52,380 in men and 29,230 in women) in 2024 [1]. Clear cell renal cell carcinoma (ccRCC) is the most common subtype of RCC, accounting for approximately 70-80% of all kidney cancer cases [1-4]. Despite recent advances in cancer treatment, the survival of ccRCC is poor, almost 30% patients with localized disease eventually develop to metastases even with early surgical treatment [3, 4]. Therefore, developing accurate prognostic biomarkers or treatment outcome prediction models is of great importance to assist in the risk stratification, treatment plan, and follow-up management of ccRCC patients.

Currently, the TNM staging system and histopathological grading are the primary methods used for cancer risk stratification and treatment planning [2-4]. However, these approaches rely on population-level data and do not account for individual tumor biology. Additionally, histopathological analysis depends on the pathologist's experience, which can introduce subjectivity. As a result, these methods may have variable accuracy in predicting patient outcomes. Recent efforts have focused on analyzing gene expression, mutations, and other molecular characteristics to better stratify patients based on their tumor biology.

With the advancement of sequencing technologies and a variety of machine learning algorithms, great efforts are ongoing to develop more comprehensive and integrated risk stratification biomarkers and models that incorporate multi-omics data, clinical information, and advanced image techniques to better predict outcomes and guide personalized treatment strategies for curing cancers [5-14]. The advent in high-throughput sequencing technologies and the establishment of large-scale cancer genomics projects, such as The Cancer Genome Atlas (TCGA), have offers unprecedented opportunities to develop robust, multi-modal predictive models that integrate diverse biological information [15]. Among these projects and datasets, the TCGA Kidney Renal Clear Cell Carcinoma (KIRC) dataset provides extensive molecular characterization, including genomic, transcriptomic, epigenomic, and proteomic data from hundreds of ccRCC patients. In addition, imaging data including CT and pathology image in the form of hematoxylin and eosin-stained (H&E) whole slide images (WSI) are also provided for part of the included patients. A lot of studies have leveraged the TCGA-KIRC dataset to identify prognostic biomarkers and develop predictive models for ccRCC [5-10]. However, most of these ccRCC studies focus on single modalities with or without combination of clinical information.

In this work, we aim to leverage the TCGA-KIRC dataset to develop and validate predictive outcome models for overall survival (OS) and disease-free survival (DFS) in ccRCC patients. By integrating multi-modal data and different machine learning and deep learning techniques, we target to enhance the predictive accuracy and robustness of ccRCC treatment outcome prediction, which could ultimately contribute to improved patient management and personalized therapy.

## 2. Materials and Methods

### 2.1 Dataset

The TCGA-KIRC dataset has been a cornerstone for renal cancer research, it includes comprehensive genomic, transcriptomic, pathology, and clinical data of 537 patients [15], which enables a multitude of studies to enhance our understanding of ccRCC. To assist the data analysis for researchers from or outside the field of bioinformatics, a set of tools were published, such as the cBioPortal [16], TCGA-Assembler [17], Firebrowse [18], and LinkedOmics [19]. These tools provide pre-preprocessed data, especially multi-omics data to help handle this vast amount of data. In this study, we focused on the treatment outcome prediction of ccRCC patients, including OS and DFS. Patients' clinical data, comprising demographic, diagnosis, treatment and follow-up data, multi-omics data, comprising pre-processed mRNA expression, miRNA expression (miRSeq), and DNA methylation data, and pathology data in the form of whole side image (WSI) are collected. The selection of modalities of multi-omics data is based on the reported prognostic performance in previous research and availability of data. The clinical data are from Genomic Data Commons Data Portal and cBioPortal, while the pre-processed multi-omics data is from TCGA Preprocessed Multi-Omics Cancer Benchmark Dataset [20]. OS is defined as the duration from initial diagnosis to last follow-up or patient death, and DFS is defined as the duration between initial diagnosis and last disease-free follow-up or cancer recurrence/progression. After excluding patients who miss any of the modalities of interest or have incomplete follow-up data of OS or DFS, we collected data of 226 ccRCC patients from TCGA-KIRC dataset in this study (67 patients have disease progression or recurrence at last follow-up and 38 patients are deceased). The clinical characteristics of the included patients are summarized in Tabel 1.

**Table 1.** Patient and tumor characteristics.

| Characteristics | | Total=226 |
|---|---|---|
| | | Number and Median [Range] or Percentage |
| Gender | Male | 146 (65.9%) |
| | Female | 80 (34.1%) |
| Age | | 60 [26~86] |
| Race | White | 186 (82.3%) |
| | African American | 38 (16.8%) |
| | Other and Unknown | 2 (0.9%) |
| Ethnicity | Hispanic or Latino | 8 (3.5%) |
| | Not Hispanic or Latino | 181 (80.1%) |
| | Unknown | 37 (16.4%) |
| Disease Laterality | Left | 105 (46.5%) |
| | Right | 121 (53.5%) |
| T Stage | T1 | 120 (53.1%) |
| | T2 | 32 (14.2%) |
| | T3 | 72 (31.9%) |
| | T4 | 2 (0.9%) |
| N Stage | NX | 122 (54.0%) |
| | N0 | 99 (43.8%) |
| | N1 | 5 (2.2%) |
| M Stage | MX | 24 (10.6%) |
| | M0 | 173 (76.6%) |
| | M1 | 27 (12.0%) |
| | Unknown | 2 (0.9%) |
| Neoplasm Histologic Grade | GX | 1 (0.4%) |
| | G1 | 7 (3.1%) |
| | G2 | 101 (44.7%) |
| | G3 | 89 (39.4%) |
| | G4 | 26 (11.5%) |
| | Unknown | 2 (0.9%) |
| Prior Cancer | Yes | 31 (13.7%) |
| | No | 195 (86.3%) |

| | |
|---|---|
| Follow Up (Month) | 33.9 [0, 149.1] |

## 2.2 Clinical and multi-Omics feature based CPH model

Cox Proportional Hazards (CPH) model is the most frequently used approach for survival analysis and survival model construction in cancer treatment outcome prediction [21]. It is a semiparametric approach that computes the impact of a set of features on the hazard (i.e., risk) of an event occurring (in this study, death and cancer recurrence). The hazard of a patient is the product of a population baseline hazard which changes over time and a time-invariant partial hazard of which the corresponding log-partial hazard (risk score) is a weighted combination of predictive features. In this work we constructed eight CPH models in total for OS and DFS prediction with clinical features, mRNA features, miRSeq features, and methylation features separately. As the multi-omics features is of very high dimension (20531, 588, and 20115, for mRNA, miRSeq, and methylation, respectively in this study), and effective feature selection is important in CPH model construction to enhance model accuracy, interpretability, and robustness for survival predictions, we used an iterative forward feature selection strategy with feature pre-selection for all the traditional CPH model training.

The workflow of the training of CPH models with clinical features and different omics features are summarized in Fig 1.a. It contains feature pre-selection and iterative forward feature while CPH model construction. At first, we exclude features which are missed for more than 20% of the included patients, and we replace the remaining missing data with the median values of the corresponding features. Categorical features are then coded with one-hot coding, numerical features are normalized with z-score normalization. An unsupervised Spearman rank correlation analysis is then performed for the original feature set, features with correlation coefficients higher than 0.8 to previous features are excluded from the feature pool. After unsupervised feature selection, 10 times of random data partition within the training cohort are conducted to split training cohort to sub-training and sub-validation data, supervised univariate CPH analysis is conducted to select predictive feature (C-index>0.5) for the following process.

The pre-selected features will be ranked according to their validation performance on the corresponding outcome, for the following iterative forward feature selection while CPH model construction. In the first iteration, the optimal feature set is initialized with the feature with the best validation performance among all the univariant models, and the first CPH model will be trained with feature(s) in current optimal feature set, the validation performance is recorded as the current best performance. Then, a set of CPH models will be trained with feature(s) in the optimal feature set and each of the remaining features in the pre-selected with pool. If the best performance of this model set is better than the current best performance, the feature which has the best validation performance when combined with current optimal feature set will be added to the optimal feature set, the current best performance will be updated, and next iteration can start. If the validation performance stops increasing or the number of iterations reaches the pre-defined maximum number (maximum number of selected features), the iteration will stop. The feature set which achieves the best validation performance during iteration will be used as the optimal feature set, and the corresponding CPH model will be stored as the trained-CPH model for the corresponding outcome and data modality. The validation performance is saved as well for calculating the weighting factor for multi-modality fusion. The details for model fusion will be introduced in a later section. We empirically set the maximum number of selected features to 20 since traditional machine learning usually suggests using one-tenth of the train sample size as the maximum feature number to balance dataset efficacy and overfitting.

### 2.3 WSI based deep CPH model

#### 2.3.1 WSI data pre-processing

WSI is usually more than 10,000×10,000 pixels, and data pre-processing including image background and hole removal, image patching, and data storage, is of great importance to conduct effective and efficient computational pathology research with this data modality. In this study, we adopted the WSI pre-processing in the CLAM (clustering-constrained attention-based multiple-instance learning) toolbox [22]. We generate a downsample of the slide and convert the color space from RGB to hue–saturation–value (HSV), then we compute binary masks for tissues using binary thresholding in the saturation channel to exclude the background and holes. Median blurring and morphological closing are used to smooth tissue contours and remove artifacts. Finally, we divide the segmented tissue regions into contiguous 512×512 patches and store them as HDF5 (Hierarchical Data Format version 5) files for feature extraction.

#### 2.3.2 Feature encoding with foundation models

Four pretrained feature encoders are collected and compared in this study for constructing WSI-based survival prediction models. These encoders include: 1) ResNet, one of the most widely used CNN model for computer vision task and was pretrained with ImageNet dataset [23]; 2) CTransPath, a CNN-transformer-hybrid unsupervised contrastive learning model for histopathological image classification, it was trained with large-scale and well-annotated datasets including TCGA and pathology AI platform [24]; 3) UNI, a general-purpose image-based foundation model for computational pathology, which was trained using more than 100 million image from 100,000 diagnostic WSIs across 20 major tissue types without TCGA data included [25]; 4) CONCH, a vision-language foundation model for computation pathology, which was trained with over 1.17 million image-caption pairs without including any TCGA data. We use these pre-trained models as encoders, and four sets of WSI features (embeddings) for each of the pre-processed WSI patches are generated and stored for model construction and encoder comparison.

#### 2.3.3 Deep CPH model

DeepSurv is a deep learning-based model designed for risk score prediction in survival outcome prediction [26]. It builds upon the principles of the CPH model but leverages the power of neural networks to capture complex, non-linear relationships between covariates and the risk score, regardless of the input data modality. We build our model with the same principle of DeepSurv and adopt the same loss function (Average Negative Log Partial Likelihood Loss) and optimization setting for model training. Our model architecture is shown in Fig 2.b. The network takes WSI embeddings as the input, each of the embeddings is firstly projected to a vector of 256 length, ReLU activation and random dropout is applied to the projected vector. Then, the vector is forward to a multi-head attention layer, where we used Nyström Attention module here as an efficient approximation technique for the self-attention calculation [27]. Average pooling is applied on the deep features from multi-head attention layer to integrate the prognostic value from each of the single WSI patches, it also enables the network to take arbitrary number of input embeddings, as the number of WSI patches varies for different patients and we don't need to use all the patches for model training. Finally, a linear fully connection layer is applied to combine the deep features to risk score for survival prediction.

### 2.4 Multi-modal ensemble model (MMEM) C-index

As the same dataset is used to construct the clinical CPH model, multi-omics CPH models, and WSI feature based deep-CPH model, the population-based baseline hazard should be the same for these models, and the risk scores of test data from these models can be weighted ensembled to achieve multi-modal fusion. The risk score $r_i$ of patient $i$ in MMEM is calculated according to formula (1), where $m$ is the index of modalities, and $M$ is the total number of modalities in MMEM ($M = 5$ in this study).

$$r_i = \sum_{m=1}^{M} r_{i,m} w_m \tag{1}$$

The weighting parameter for modality $m$ is determined by the following formula, where $p_m^{val}$ is the validation performance of model built with modality $m$. We use OS and DFS prediction C-indexes as the performance evaluation metrics here.

$$w_m = p_m^{val} / \sum_{m=1}^{M} p_m^{val} \tag{2}$$

### 2.5 Performance evaluation

Five-fold cross-validation is performed to train and evaluate our method. All the patients are randomly split into 1 of 5 folds, and the training and test cohorts follow a 4:1 fashion. Testing cohort prediction results are combined for model evaluation.

We utilize the C-index to evaluate the model's capability to predict a higher risk of events for patients at an earlier time while taking data censoring into consideration. The 95% confidence interval (CI) of C-index is calculated through 1000-time bootstrapping. Unpaired t-test is used to compare the distributions of bootstrapped C-index from different models. We use median risk values of OS and DFS as the thresholds to divide patients into high- and low-risk groups of the corresponding outcomes. We then compare their Kaplan-Meier survival curves using log-rank test to evaluate the effectiveness of risk stratification. To assess the model's ability to identify high-risk patients for death and disease recurrence at 1, 3, and 5 years, we employ the area under the receiver operating characteristic curve (AUROC) analysis. Patients who were censored before 1, 3, and 5 years were excluded for the corresponding AUROC measurement. The 95% CI of AUCs are estimated via Delong's method [28], and Delong's test is used to measure the difference in ROC curves. A p-value of ≤ 0.05 is considered significant for all the statistic tests.

### 2.6 Implementation details

Our experiments are implemented using Python (version 3.10) environment with PyTorch (version 1.13) framework on a 24GB NVIDIA GTX3090 GPU. The WSI images are cropped into 512 by 512 patches and then downsampled to size of 224 by 224 for feature extraction with all different models. During the training process of the deep CPH model in MMEM, 4096 WSI patches are randomly selected for prediction, and all the WSI patches are used for model testing. We use Adam as the optimizer, batch size of 1, and epoch number of 100 for training all the deep CPH models. The learning rate is set as 0.001 with dynamic learning rate reducing when the loss value on validation cohort stops decreasing for 5 epochs (gamma=0.1).

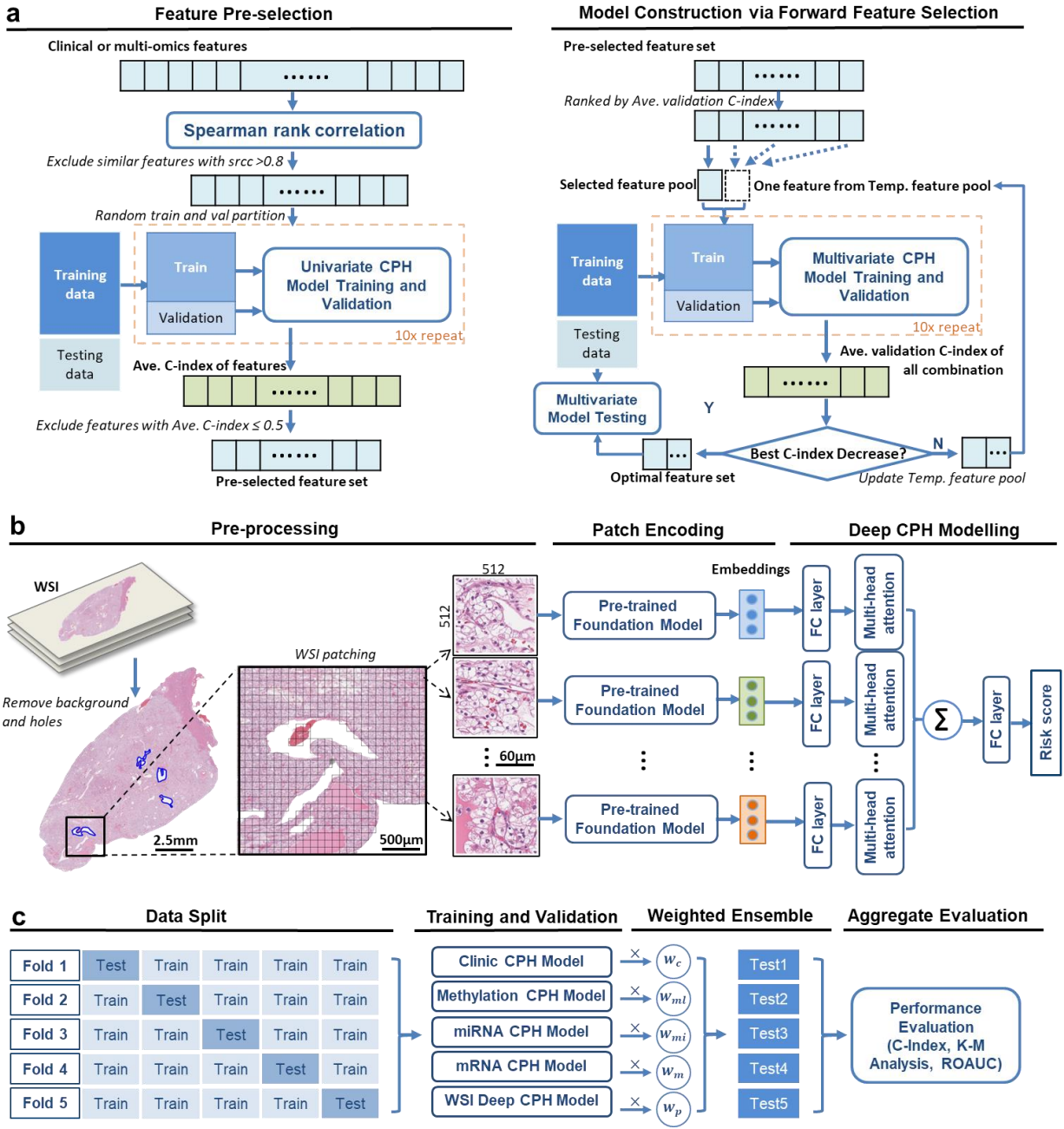

**Fig 1.** Overview of multi-modal ensemble model (MMEM) for clear cell renal cell carcinoma (ccRCC) treatment outcome prediction. **a**) Feature pre-selection and Cox Proportional Hazard (CPH) model construction using iterative forward feature selection for clinical and multi-omics data. **b**) Whole slide image pre-processing, deep feature generation, and deep CPH model construction for pathology data. FC layer stands for fully connection layer. **c**) Five-fold cross validation, weighted model fusion, and survival prediction performance evaluation. The weighting factor for each modality modal is based on the corresponding average validation concurrence-index (c-index). C-index, Kaplan–Meier (K-M) analysis, and the area under the receiver operating characteristic curve (AUROC) analysis of the aggregated test set (full dataset) are used for model evaluation.

## 3. Results

### 3.1 Clinical and multi-omics CPH model

With the feature pre-selection process, we downsized the dimension of clinical features from 17 to 11 and 14 for OS and DFS predictions respectively, mRNA features from 20531 to around 2000 for different outcomes and training cohorts, miRSeq features from 588 to around 230 for different outcomes and training cohorts, and DNA methylation features from 20116 to around 2000 for different outcomes and training cohorts. Of note, the number of pre-selected feature is unstable for multi-omics data, since the original data is of high dimension and the selected feature is sensitive to data partition. The OS and DFS prediction performances are summarized in Table 2, while the binary outcome prediction performances using risk score at different follow-up time points are summarized in Table 3. The corresponding K-M curves for different risk groups and ROC curves are presented in Fig 2-3. The selected features for different modalities are summarized in the Supplementary Material.

**Table 2.** Concordance index and its 95% confidence interval in prediction overall survival (OS), and disease-free survival (DFS) using different models and data modalities (* means the C-index of this model performs significantly higher than others, bolded value is the best performance for the corresponding outcome).

| Model | C-index (95% CI) | |
|---|---|---|
| | OS | DFS |
| Clinical CPH | 0.785 (0.671, 0.888) | 0.803 (0.724, 0.876) |
| mRNA CPH | 0.663 (0.546, 0.775) | 0.646 (0.552, 0.740) |
| miRSeq CPH | 0.590 (0.472, 0.705) | 0.677 (0.587, 0.765) |
| DNA Methylation CPH | 0.687 (0.562, 0.798) | 0.681 (0.599, 0.759) |
| WSI Deep CPH | 0.734 (0.644, 0.824) | 0.739 (0.666, 0.807) |
| MMEM | **0.820 (0.721, 0.908)**$^*$ | **0.833 (0.773, 0.891)**$^*$ |
| MMEM (Uniform Weight) | 0.802 (0.714, 0.886) | 0.816 (0.755, 0.875) |

**Table 3.** Area under the receiver operating characteristic curve (AUROC) and its 95% confidence interval in prediction patient death and disease recurrence within 1, 3, and 5 years after diagnosis using different models (bolded value is the best performance for the corresponding outcome).

| Model | Outcome | AUC (95% CI) | | |
|---|---|---|---|---|
| | | 1-year | 3-year | 5-year |
| Clinic CPH | Patient Death | 0.827 (0.697, 0.957) | 0.781 (0.675, 0.886) | 0.805 (0.707, 0.903) |
| | Disease Recurrence | 0.818 (0.724, 0.912) | 0.821 (0.739, 0.903) | 0.830 (0.752, 0.908) |
| mRNA CPH | Patient Death | 0.723 (0.573, 0.872) | 0.665 (0.556, 0.773) | 0.684 (0.574, 0.794) |
| | Disease Recurrence | 0.630 (0.520, 0.740) | 0.685 (0.565, 0.752) | 0.666 (0.563, 0.768) |
| miRSeq CPH | Patient Death | 0.542 (0.373, 0.711) | 0.622 (0.509, 0.734) | 0.615 (0.500, 0.730) |
| | Disease Recurrence | 0.684 (0.581, 0.787) | 0.726 (0.639, 0.814) | 0.704 (0.605, 0.803) |
| DNA Methylation CPH | Patient Death | 0.765 (0.593, 0.937) | 0.708 (0.591, 0.826) | 0.693 (0.575, 0.810) |
| | Disease Recurrence | 0.666 (0.565, 0.767) | 0.696 (0.609, 0.784) | 0.675 (0.574, 0.775) |
| WSI Deep CPH | Patient Death | 0.714 (0.605, 0.822) | 0.753 (0.655, 0.852) | 0.746 (0.649, 0.846) |
| | Disease Recurrence | 0.742 (0.658, 0.827) | 0.773 (0.696, 0.851) | 0.704 (0.606, 0.802) |
| MMEM | Patient Death | **0.848 (0.735, 0.960)** | **0.831 (0.738, 0.925)** | **0.846 (0.757, 0.935)** |
| | Disease Recurrence | **0.850 (0.780, 0.920)** | **0.862 (0.795, 0.925)** | **0.855 (0.785, 0.925)** |
| MMEM (Uniform Weight) | Patient Death | 0.832 (0.733, 0.930) | 0.824 (0.733, 0.914) | 0.828 (0.739, 0.917) |
| | Disease Recurrence | 0.817 (0.747, 0.887) | 0.851 (0.789, 0.913) | 0.826 (0.747, 0.904) |

### 3.2 Comparison of different feature encoder in WSI based deep CPH model

Survival prediction and binary outcome prediction performances when using WSI features from different encoders are summarized in Table 4. WSI features encoded by UNI have the best performance among these four encoders, which achieve 0.734 and 0.739 C-indexes for OS and DFS predictions, and 0.753 and 0.773 for binary patient death and cancer recurrence predictions before follow-up of 3 year. CONCH and CTransPath have similar performance on OS and patient death prediction, but CONCH has better DFS and cancer recurrence prediction performance. Among all the encoders, ResNet, which is not pre-trained with pathology data, has the lowest performance for all tasks.

**Table 4.** Concordance index and area under the receiver operating characteristic curve (AUROC) analysis for whole slide image (WSI) based deep CPH models built with feature embedding from different baseline models (bolded value is the best performance for the corresponding outcome).

| Feature Encoder | C-index (95% CI) | | AUROC (3-Year, 95% CI) | |
|---|---|---|---|---|
| | OS | DFS | Patient Death | Cancer Recurrence |
| ResNet | 0.603 (0.512, 0.693) | 0.616 (0.545, 0.711) | 0.612 (0.504, 0.724) | 0.641 (0.562, 0.770) |
| CTransPath | 0.722 (0.623, 0.814) | 0.691 (0.597, 0.802) | 0.745 (0.655, 0.839) | 0.712 (0.619, 0.822) |
| UNI | **0.734 (0.644, 0.824)** | **0.739 (0.666, 0.807)** | **0.753 (0.655, 0.852)** | **0.773 (0.696, 0.851)** |
| CONCH | 0.719 (0.609, 0.820) | 0.720 (0.645, 0.799) | 0.743 (0.642, 0.821) | 0.763 (0.621, 0.874) |

### 3.3 Performance of MMEM

The performance of MMEM which weighted fused the predicted risk scores from clinical model, multi-omics models, and WSI model, is summarized in Tables 2-3, and Fig 2-3. MMEM achieves 0.820 and 0.833 C-indexes for OS and DFS predictions, and 0.831 and 0.862 for binary patient death and cancer recurrence predictions before follow-up of 3 year, which outperforms single-modality models on all tasks and improved the C-index by 4.5% and 3.7% compared to the best single-modality model (clinical CPH model) on OS and DFS predictions, separately. Among all the single-modality models, clinical model has the best performance and highest weight in modality fusion, and WSI model has the second highest predictive ability. In terms of K-M cures for different risk groups and outcomes, MMEM has the best risk group partition according to log-rank test (smallest p-value), which is consistent with the survival prediction performance and binary outcome prediction performance. The validation performance of single-modality models for calculating the weighting factors are summarized in Supplementary Material. Among them, the clinical feature based CPH model contributes the most to MMEM. When we switch the modality weighting scheme from validation performance based weighted fusion to uniform weighted fusion, the performance on both survival prediction and binary prediction decreased (Tables 2-3, MMEM Uniform Weight), however the performance is still higher than single-modality model, which demonstrates the robustness of the proposed method.

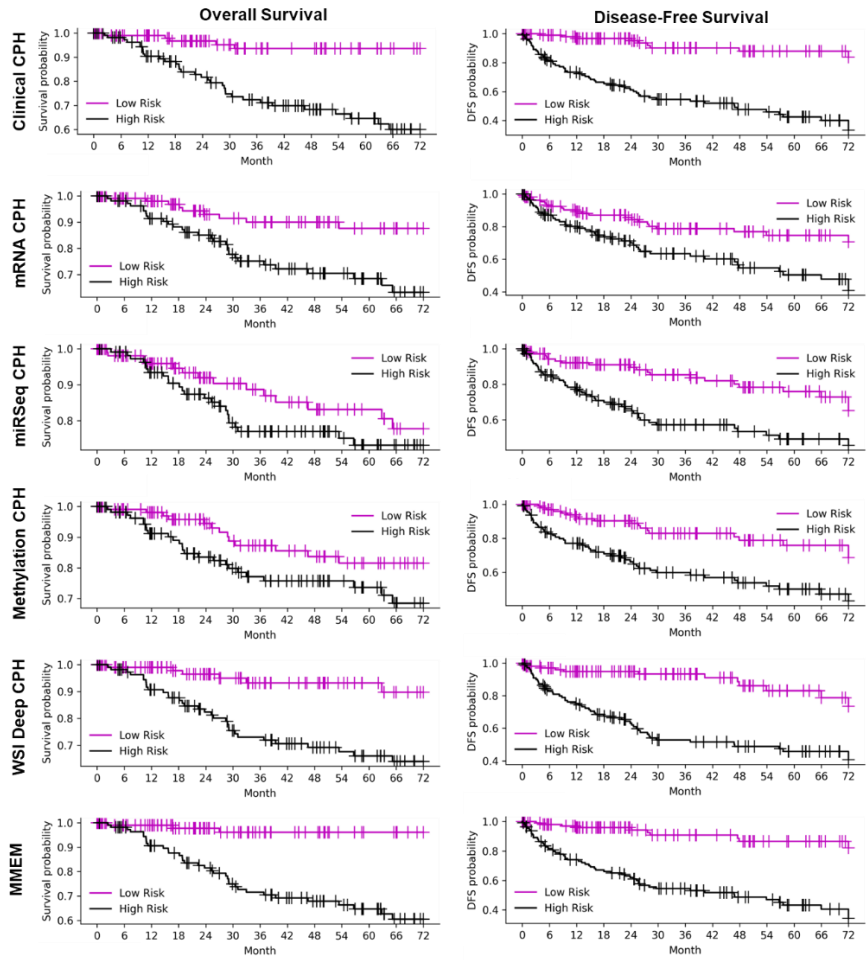

**Fig 2.** Kaplan-Meier curves of low-risk and high-risk groups for different outcomes identified with the risk score from different models and modalities. Median risk score is used as the threshold for risk partition. P-values are less than 0.01 for all the risk group pairs except for methylation CPH model for OS prediction (p=0.13).

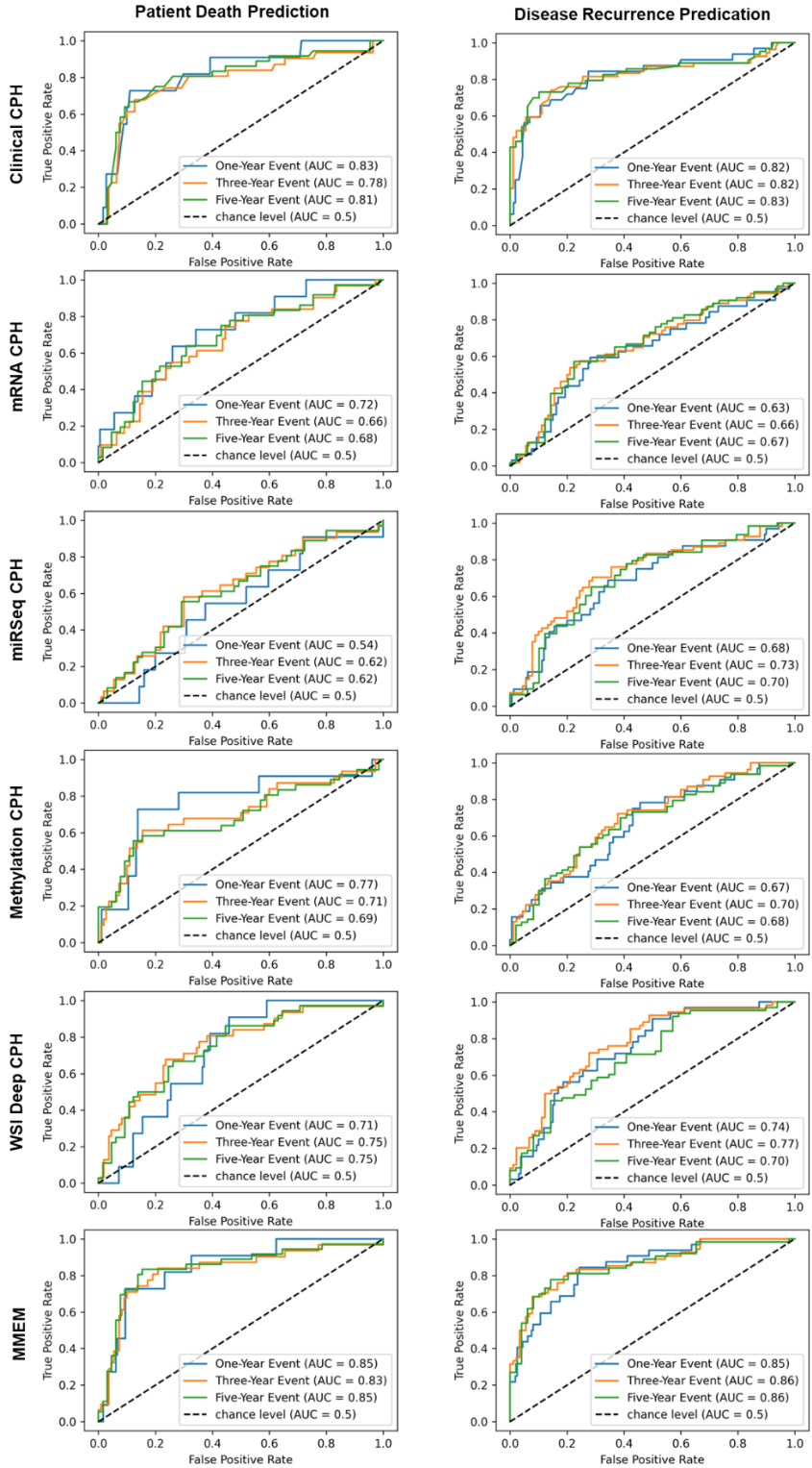

**Fig 3.** Receiver operating characteristic curves (ROCs) for patient death prediction and disease recurrence prediction using models trained with different data modalities and multi-modal ensemble model (MMEM). Follow up of 1, 3 and 5 years are selected as prediction time points.

## 4. Discussion

Our study utilized the TCGA-KIRC dataset to develop predictive models for OS and DFS in patients with ccRCC. The integration of multi-modal data, combining gene expression, DNA methylation, miRNA expression, histopathology image, and clinical information, provided a comprehensive and accurate assessment of patient prognosis. In survival prediction, the proposed MMEM achieved 0.820 and 0.833 C-indexes for OS and DFS predictions, respectively. In risk score based binary outcome prediction, MMEM got 0.848, 0.831, and 0.846 for patient death prediction at 1-, 3-, and 5-years follow-up, respectively, and 0.850, 0.862, and 0.855 for cancer progression/recurrence prediction at 1-, 3-, and 5-years follow-up, respectively. All of them outperformed any of the corresponding single-modality models.

Previous studies have highlighted the utility of the TCGA-KIRC dataset in identifying different prognostic biomarkers and constructing various outcome prediction models for ccRCC [5-10]. Wang et al. identified eight autophagy-related genes (ARGs) based on the analysis of 216 TCGA-KIRC patients' data, and their multivariate Cox analysis which integrates the ARG signature and clinicopathological information achieved C-index of 0.75 and AUC $\geq$ 0.7 for OS and patient death prediction [5]. Hu et al. identified seven methylated differentially expressed genes (MDEGs) for patient death prediction via a combined analysis of DNA methylation data and gene expression data. They built a multi-variant model with the identified biomarkers for TCGA-KIRC patient death prediction (526 patients) and achieved AUROCs of 0.735, 0.702, and 0.735, for 1-, 3-, and 5-years follow-up, respectively [6]. Wessels et al. constructed a CNN model which extracts WSI image features for 5-year patient death prediction with data from 254 TCGA-KIRC patients. The prediction AUROC is 0.75 when WSI features were used alone, and it improved to 0.81 when the CNN-based prediction results are integrated with the predictions from a multivariable clinical feature model [7]. Our findings align with these studies, confirming the prognostic value of features from different modalities. Additionally, our use of ensemble learning techniques has shown improved prediction accuracy on both OS and DFS prediction. To the best of our knowledge, as the prognostic value of five different modalities were explored for two different outcomes, our work is one of the most comprehensive studies using TCGA-KIRC dataset. Despite the difference in data selection and evaluation methodology between our work and previous studies, the positive results on both outcome prediction tasks demonstrate that there obviously is significant prognostic information encoded in the fusion of multi-modality predictions that it can capturing diverse aspects of ccRCC.

The development and implementation of general-purpose foundation models are hot topics recently in the field of computational pathology [22, 25, 29]. Foundation models are trained on massive and diverse datasets, and they are designed to handle multiple tasks simultaneously, such as classification, segmentation, and feature extraction. These models can be easily adapted to different tasks in computational pathology, with or without fine-tuning on smaller, task-specific datasets, which significantly reduces the need for large, domain-specific training datasets and the workload for feature engineering. In our work, we explored four different feature extractors for WSI data analysis, and two of them are foundation models. Features from the image base foundation model, UNI, have the best prediction accuracy (0.734 and 0.739 in OS and DFS prediction, respectively). The CNN-transformer hybrid model, CTransPath, though included TCGA-KIRC in its training data for classification model construction, performed inferior to UNI in our study. This comparison demonstrated the robustness and effectiveness of the general-purpose foundation models for computational pathology analysis which are worth further investigation with larger dataset and various diseases and outcome of interests.

Early fusion (pre-training fusion) and later fusion (post-train fusion) are two of the main methods for constructing multi-modal prediction models. However, multi-modal deep learning models need to integrate various types of data (e.g., genomic, transcriptomic, imaging, and clinical data) simultaneously, leading to an extremely high-dimensional input space. This increased complexity can make the model more prone to overfitting, especially if the dataset is not sufficiently large. Considering the risk of overfitting of current dataset and the computational complexity when integrating five different modalities, we chose later fusion to build our framework. This strategy allows us to train each single-modal model independently, enables easier scaling and parallel processing. Although we excluded patients who have missing data modality, the proposed MMEM actually can be trained with all the available data for each modality and perform individual-specific multi-modal fusion (Supplementary Material). Later fusion method also enables the use of different architectures and hyperparameters tailored to each modality to optimize their performance, such as our WSI deep CPH model. With our single-modality model trained, average fusion and weighted fusion in MMEM are evaluated, and both showed improved prediction accuracy, which again highlights the advantage of multi-modal learning in integrating the prognostic information from diverse aspects of the disease. In our experiment, the later one which gives higher weight to the modalities who have better prognostic performance has the most supervisor survival prediction accuracy. Despite the promising performance of the proposed fusion method, more advanced strategies to fuse the prediction results from each single-modality model need to be explored in our future work.

The present work has several potential limitations that require further consideration. Firstly, while our survival prediction models performed well on the TCGA-KIRC dataset, external validation is of great importance to confirm their applicability to other patient populations in our future research. Secondly, although 5 different data modalities are used for the construction MMEM, more data modalities such as diagnosis CT and SNP6 copy number data are available for TCGA-KIRC patient now. The analysis with these data modalities and evaluation of the robustness of the MMEM framework when incorporating more data modalities needs to be further investigated. In addition, we used post-training modality fusion in MMEM, the performance comparison to pre-training modality fusion technique needs to be conducted when there is a larger data cohort available. Finally, for the WSI-based deep prediction model, the interpretability analysis is essential for its clinical adoption. Techniques such as deep attention visualization need to be explored to explain deep survival models.

## 5. Conclusion

In conclusion, we proposed and validated a multi-modal ensemble approach for ccRCC treatment outcome prediction based on a comprehensive publicly available dataset TCGA-KIRC. Our findings underscore the effectiveness of integrating clinical, multi-omics, and pathology data for predicting OS and DFS in ccRCC patients via post-training modality fusion. And large-scale image based general-purpose pre-trained foundation model UNI was proved to be effective in WSI feature encoding for ccRCC outcome prediction modelling. Overall, the proposed model has the potential to contribute to personalized prognostication and guide therapeutic decision-making for management of ccRCC patients.